\documentclass[10pt,twocolumn,letterpaper]{article}

\usepackage[pagenumbers]{cvpr}      %

\usepackage{booktabs}
\usepackage{array}
\usepackage{amsmath}
\usepackage{amssymb}
\definecolor{cvprblue}{rgb}{0.21,0.49,0.74}
\usepackage[pagebackref,breaklinks,colorlinks,allcolors=cvprblue]{hyperref}

\def\confName{CVPR}
\def\confYear{2027}

\DeclareMathOperator*{\argmax}{arg\,max}

\title{ForeSay: Reading World-Model Predictions with Language}

\author{Yanshen Xie$^{1*}$\quad Shicheng Yin$^{1*}$\quad Kaixuan Yin$^{1}$\quad Yang Liu$^{1,2\dagger}$\\
$^{1}$Sun Yat-sen University\quad $^{2}$X-Era AI Lab\\
{\small \{xieysh33, yinshch5, yinkx3\}@mail2.sysu.edu.cn,\ \ liuy856@mail.sysu.edu.cn}\\
{\small Code: \href{https://github.com/HCPLab-SYSU/ForeSay}{\textcolor{black}{github.com/HCPLab-SYSU/ForeSay}}}}

\makeatletter\@namedef{ver@eso-pic.sty}{2020/10/14}\makeatother
\usepackage{pdfpages}
\begin{document}
\twocolumn[{%
\renewcommand\twocolumn[1][]{#1}%
\maketitle
\begin{center}
\captionsetup{type=figure}
\includegraphics[width=\textwidth]{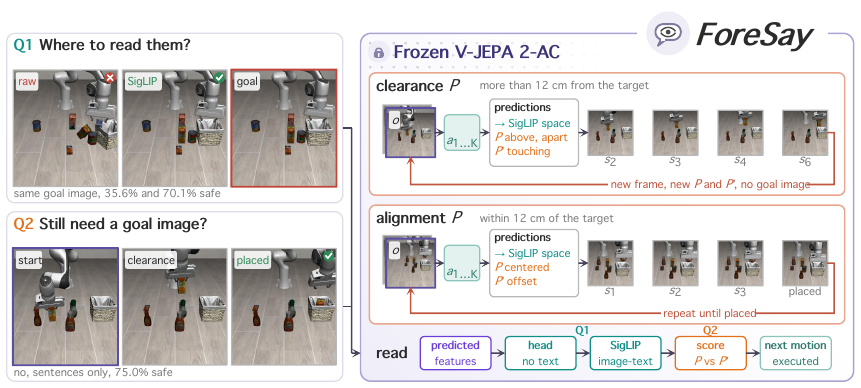}
\caption{Reading the predictions of a frozen V-JEPA 2-AC predictor. With one goal image, raw features and projection into SigLIP's space give 35.6\% and 70.1\% live safe placement (Q1). VLM-written sentences give 75.0\% without a goal image, with both geometric points estimated from RGB-D (Q2). On the right, the far- and near-phase prompts are shown from two live trials.}
\label{fig:teaser}
\end{center}
\vspace{0.6em}
}]

{\let\thefootnote\relax\footnotetext{$^{*}$Equal contribution.\quad $^{\dagger}$Corresponding author.}}
\begin{abstract}
Goal-image planners built on action-conditioned world models score each predicted outcome by its feature distance to a goal image. For a frozen V-JEPA 2-AC predictor, we ask what changes when its predicted features are compared with the same goal image in another space, and whether sentences can then take the place of goal images. To answer both questions, we map these features into SigLIP's image-text space with ForeSay, a head trained on demonstration frames without text or task-outcome labels. In rearranged LIBERO layouts, a held object must go into a basket without displacing the basket or other objects, and every planner receives a release point above the basket. A planner guided only by distance to this point places no object safely. With the same final goal image, safe placement rises from 35.6\% in raw features to 70.1\% after projection. On raw features, changing the distance metric or pooling does not close this gap. Raw distance restricted to obstacle patches recovers contact information that whole-frame distance dilutes. In the projected space, scoring the predictions with phrases fixed before testing separates the basket relations reached by candidates executed from new states. Target sentences written by a VLM from the current frame reach 84.1\% safe placement without any goal image. Staged goal images, which require a prior execution in the same layout, reach 86.4\%.
\end{abstract}

\section{Introduction}
\label{sec:intro}
Action-conditioned visual world models such as V-JEPA 2-AC and DINO-WM predict the outcome of each candidate action in a visual feature space~\cite{assran2025vjepa2,zhou2025dinowm}. Their goal-image planners score each predicted outcome by its feature distance to a goal image. The goal image specifies where the manipulated object should end and how the rest of the scene should look. In a distance over V-JEPA 2-AC's 256 feature patches, an object that covers a few of them contributes little. When such a planner disturbs the scene, it is therefore unclear whether the predictions miss the disturbance or the distance hides it.

We examine this question on placement in rearranged LIBERO-Object scenes~\cite{liu2023libero}. The robot has to move an already-held object into a basket without displacing the basket or other objects, which lie on the route in most scenes. Such a displacement is an event, and safe placement is completion without an event. Every planner we compare receives the same release point above the basket, a point that sets progress, the planning phase and the moment of release. Progress is the distance from the held object to that point and does not depend on the other objects. Guided by progress alone, the planner gives no safe placement in 264 trials under replay. With the same candidates, controller and release rule, scoring the predictions with sentences stored during live runs gives 214 safe placements in these trials.

The two runs differ only in the readout, the score given to each predicted outcome. We ask two questions about the readout. What changes when the predicted features are compared with the same goal image in another space? In that space, can sentences take the place of the goal image?

We fine-tune V-JEPA 2-AC on demonstrations and freeze it. A two-layer head, which we call ForeSay, maps its observed and predicted features to SigLIP image embeddings~\cite{zhai2023siglip} (\cref{fig:teaser}). The head is trained on demonstration frames, without text or task-outcome labels. A prediction can then be compared with an image embedding or with a sentence embedding. \Cref{tab:protocol} maps each claim below to its measurement.

\emph{The same goal image reads differently in the two spaces.} With the goal pixels, the predictor and the release point fixed, live safe placement rises from 35.6\% in raw features to 70.1\% after projection. The size of this change depends on the head. Three heads trained from different seeds give 137 to 202 of 264 safe placements under replay when the planning weight is retuned for each head, and raw features give 97. Changing the raw metric to cosine or squared distance, or pooling the raw features, gives at most 99. Executing candidates from restored states shows where the raw score loses information. Route obstacles cover about 7 of the 256 patches, and raw distance restricted to those patches ranks contact better than whole-frame distance.

\emph{In the projected space, sentences can take the place of the goal image.} Before testing, we choose one pair of phrases using real frames from other layouts and lock its wording. On 129 new states, scoring the predictions with these phrases separates the basket relations that executed candidates reach. In closed loop, a VLM reads the current frame once per decision. It writes a target sentence describing the desired next frame and a contrast, and this one response is used to score all candidate predictions at that decision. Live, sentences reach 84.1\% safe placement without any goal image. Staged goal images show a mid-execution frame while the held object is far from the basket and the final frame once it is near. Taken from a prior successful execution in the same layout, they reach 86.4\%. Under replay, staged frames from the demonstration each layout was built from require no such execution and give 75.8\%, and sentences give 81.1\%. When the same VLM chooses motions directly from the current frame and the same release point, live safe placement is 38.3\%.

\section{Related Work}
\label{sec:related}

\paragraph{Prediction and goal spaces.}
World models support planning with rewards~\cite{ha2018worldmodels,hafner2019planet,hafner2020dreamer,hafner2025dreamerv3,chua2018pets,hansen2022tdmpc,hansen2024tdmpc2}, designated-pixel goals~\cite{finn2017foresight} or goal-image distances~\cite{ebert2018foresight,nair2018rig}. Joint-embedding predictive architectures~\cite{assran2023ijepa,bardes2024vjepa} predict in feature space, and DINO-WM, PLDM, V-JEPA 2-AC and later work plan in visual feature spaces~\cite{zhou2025dinowm,sobal2025pldm,assran2025vjepa2,terver2025jepawms,zhang2026hwm,goswami2025dexwm}. CLIP and SigLIP provide an image-text space~\cite{radford2021clip,zhai2023siglip}, and pretrained vision-language models supply rewards and success signals~\cite{du2023success,rocamonde2024vlmrm,sontakke2023roboclip,ma2023liv}. LiT aligns text to a frozen image encoder~\cite{zhai2022lit}, and Text-to-Concept linearly maps vision-only features to the CLIP embeddings of the same images~\cite{moayeri2023text2concept}. ForeSay applies such an alignment to action-conditioned predictions, whose readouts we compare in closed loop.

\paragraph{Training language-readable predictions.}
Grounded World Model trains an action-conditioned predictor to forecast frozen video-language embeddings without language labels~\cite{li2026gwm}. Its action inputs are rendered robot images, and instructions score predicted trajectories. ForeSay is a head on a frozen visual predictor, so the raw predictions remain available for readout comparisons. LOReL learns language-conditioned rewards~\cite{nair2021lorel}, LAGO learns latent subgoals from language-labeled demonstrations~\cite{barbeau2026lago}, and LeVLWM pretrains a world model and text encoder together~\cite{hua2026levlwm}. Our head uses demonstration images and pretrained SigLIP alignment. Language also enters model and policy learning in Dynalang and GenRL~\cite{lin2024dynalang,mazzaglia2024genrl}, and video generation in UniPi and video language planning~\cite{du2023unipi,du2024vlp}.

\paragraph{VLM access to candidates.}
VLMPC tracks VLM-detected objects through predicted videos~\cite{zhao2024vlmpc}. FOREWARN fine-tunes a VLM to describe and rank latent outcomes~\cite{wu2025forewarn}, and Semantic World Models query action-conditioned futures~\cite{berg2025swm}. Our planner queries the VLM once per decision on the current frame and scores all candidates through SigLIP. Approaches that work on the observed scene construct value maps in VoxPoser~\cite{huang2023voxposer}, keypoint constraints in ReKep~\cite{huang2024rekep}, or marked points and action proposals in MOKA and PIVOT~\cite{fang2024moka,nasiriany2024pivot}.

\section{Method}
\label{sec:method}

\begin{figure*}[t]
\centering
\includegraphics[width=\textwidth]{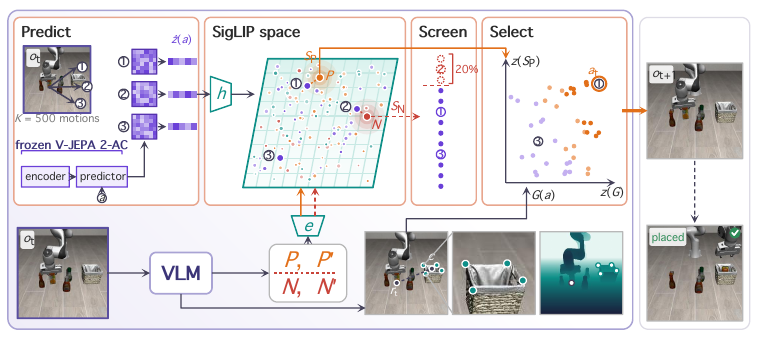}
\caption{One planning step with the sentence readout. ForeSay, a head $h$ trained without text or task-outcome labels, maps the predictions for $K=500$ sampled motions into SigLIP's space, where the text encoder $e$ embeds the VLM's target pair $(P,P')$ and screening pair $(N,N')$. The simulator or VLM-marked points with depth give $c$ and $r_t$. The SigLIP plane, feature grids and score plots are schematic.}
\label{fig:overview}
\end{figure*}

\Cref{fig:overview} shows the planning cycle within which we compare readouts. The predictor supplies candidate-dependent features. A readout scores each predicted outcome, and a geometric term measures approach to the destination. Most comparisons in \cref{sec:exp} change only the readout, its goal input and its weight. \Cref{tab:position} lists the information each component uses.

\begin{table}[t]
\centering
\caption{Information used in our planner and in the direct-control reference. The predictor and head are trained without text or task-outcome labels. SigLIP supplies pretrained image-text alignment. Geometry comes from the simulator in readout comparisons and from RGB-D point estimates in the perception experiment.}
\label{tab:position}
\footnotesize
\setlength{\tabcolsep}{3pt}
\begin{tabular}{@{}p{0.3\linewidth}p{0.65\linewidth}@{}}
\toprule
Component & Information supplied \\
\midrule
Predictor fine-tuning & Demonstration frames, robot states and actions \\
Head training & Observed and predicted features paired with SigLIP image embeddings \\
Progress and release & Robot state, release point and held-object reference offset \\
Image readout & Final or staged goal images \\
Sentence query & Current RGB frame, task, declared names and distance-based phase \\
Direct VLM control & Current RGB frame, declared names, end-effector position and release point \\
\bottomrule
\end{tabular}
\end{table}

\subsection{Problem setting}
\label{sec:setting}
The robot moves an already-held object into a basket. At step $t$, it observes one RGB frame from a fixed camera and the robot state, then executes an end-effector translation of at most 7.5\,cm. The geometric destination $c$ is a release point above the basket opening, and the reference point $r_t$ lies at the bottom of the held object. Their distance sets geometric progress, the phase and the release condition. The point $c$ fixes where the object should end. Whether a motion keeps the basket and the other objects in place enters only through scores computed from the predictions. With $c$ given to every planner, the comparisons measure what a readout adds once the destination is known. We vary the target score and keep screening fixed.

For controlled readout comparisons, the simulator supplies $c$ and the initial reference-point offset. In the perceived variant, a VLM marks four points on the basket rim and the lowest visible point of the held object, and depth maps these marks to 3D, with the target above the rim center (supplementary Sec.~A). Both variants use the same readout and selection equations.

\subsection{The predictions under study}
\label{sec:wm}
V-JEPA 2-AC pairs a pretrained visual encoder with an action-conditioned predictor~\cite{assran2025vjepa2}. We fine-tune the released predictor on LIBERO-Object demonstrations with the official training code, then freeze the entire world model (settings in supplementary Sec.~J). Demonstrations that provide evaluation starts are excluded from this training. Given current image features, robot state and a candidate translation, the predictor outputs future patch features. All readouts share this predictor, so a change of readout leaves the predicted features for a given state and action fixed.

\subsection{Aligning predictions and reading them}
\label{sec:head}
Let $\hat Z_j(a)$ be the predicted feature at patch $j$ for candidate $a$, with $M$ patches of dimension $d$, and let $\hat z(a)$ be their mean. A two-layer MLP $h$ maps this 1408-dimensional vector through 2048 hidden units with GELU and LayerNorm into the image-text space of SigLIP So400m/14 at 384 pixels~\cite{zhai2023siglip}. Training uses two kinds of image pairs. Observed encoder features pair with SigLIP embeddings of their frames. Predictions under recorded actions pair with embeddings of the frames those actions reached. InfoNCE~\cite{oord2018cpc} pulls each mapped vector toward its paired image embedding and contrasts other images in the batch. We use batch size 256, temperature 0.05 and learning rate $10^{-3}$ for 50 epochs. Neither text nor collision or success labels enter head training, and the predictor receives no update. One head serves all candidates, sentences and layouts of an evaluation run.

The raw readout $R_{\mathrm{raw}}(a,g)$ is the negative mean L1 distance between the predicted patch features $\hat Z_j(a)$ and the goal image's encoder features $Z_j(g)$, averaged over patches and dimensions. This is the score with which V-JEPA 2-AC plans toward a goal image, and we use it as the reference readout. The projected readouts use the unit-normalized head output $v(a)$ and unit SigLIP embeddings $u_I$ of images and $u_T$ of sentences. A goal image $g$ gives the cosine $R_{\mathrm{proj}}(a,g)=v(a)^\top u_I(g)$, and a sentence $s$ with contrast $s'$ gives
\begin{equation}
S(a,s,s')=v(a)^\top\big(u_T(s)-u_T(s')\big).
\label{eq:readout}
\end{equation}
Each image or sentence query defines one direction shared by all candidates at a decision, and the predicted outcome supplies the candidate-dependent vector.

\subsection{Specifying the successor relation}
\label{sec:queries}
At each decision, the VLM receives the current frame, task, declared object names and phase. The phase is far when $\lVert r_t-c\rVert_2>12$\,cm and near otherwise. The VLM writes a target sentence $P$ describing the desired next frame and a contrast $P'$. In the far phase the prompt asks for the held object separated from nearby objects and near the basket, and in the near phase for the object centered in the receiving region. Sentences state static visible relations, avoid motion verbs and coordinates, and have at most 30 words.

The same response contains a contact-related sentence $N$ and its contrast $N'$ for screening. Define $S_P(a)=S(a,P,P')$ and $S_N(a)=S(a,N,N')$. The VLM sees only the current observation, and all candidates are scored with the same returned pairs. Full prompts appear in supplementary Sec.~H.

\subsection{A planner shared by all readouts}
\label{sec:select}
The planner samples $K=500$ translations, comprising zero motion, the straight motion toward $c$ and 498 Gaussian samples with standard deviation 7.5\,cm per axis. Motions longer than 7.5\,cm are shortened. Candidates below 5\,mm are ineligible unless $r_t$ is within 2\,cm of $c$. Screening removes the 20\% of eligible candidates with the highest $S_N$. Among the remainder $\mathcal A_t$, the planner executes
\begin{equation}
a_t=\argmax_{a\in\mathcal A_t}\; z\big(G(a)\big)+w\,z\big(S_P(a)\big),
\label{eq:select}
\end{equation}
with $w=0.6$, progress $G(a)=-\lVert r_t+a-c\rVert_2$, and $z$ standardizing each score over $\mathcal A_t$. Image-goal planners replace $S_P$ by $R_{\mathrm{raw}}$ or $R_{\mathrm{proj}}$ and select their own $w$ on development layouts. The sampler, progress term, screening and release rule are shared.

An operational-space controller~\cite{khatib1987osc} executes the chosen translation to within 3\,mm or for at most 40 control frames. The planner observes the resulting frame and replans. It opens the gripper within 1\,cm of $c$ and permits at most 16 decisions.

A live trial averages 8.6 decisions and queries the VLM once per decision, so one response scores all 500 candidates of that decision.

\section{Experiments}
\label{sec:exp}
\Cref{sec:progress} measures what the shared release point achieves by itself. \Cref{sec:controlled,sec:space} answer the first question, and \cref{sec:probe,sec:main,sec:ablation} the second.

\begin{table}[t]
\centering
\caption{Evidence map. Closed-loop cells use the 88 evaluation layouts with seeds 1 to 3 (264 trials) unless marked dev., which uses 15 development layouts. Weights are selected on development layouts with seeds 0, 4 and 5. Probe states come from evaluation layouts.}
\label{tab:protocol}
\footnotesize
\setlength{\tabcolsep}{2.3pt}
\begin{tabular}{@{}>{\raggedright\arraybackslash}p{0.44\linewidth}>{\raggedright\arraybackslash}p{0.29\linewidth}>{\raggedright\arraybackslash}p{0.23\linewidth}@{}}
\toprule
Claim & Evidence & Protocol \\
\midrule
\multicolumn{3}{@{}l}{\emph{Shared release point}} \\
Progress alone places nothing safely & \Cref{tab:controlled} & Replay \\
\midrule
\multicolumn{3}{@{}l}{\emph{Question 1, the same goal image in two spaces}} \\
Projection changes outcomes & \Cref{tab:controlled} & Live, replay \\
Effect size across heads & \Cref{tab:heads} & Replay \\
Weight, metric and pooling & \Cref{tab:outcomes} & Replay, dev. \\
Raw distance dilutes local cues & Obstacle-patch probe & 33 states \\
\midrule
\multicolumn{3}{@{}l}{\emph{Question 2, sentences in the projected space}} \\
Scores track executed relations & Locked and generated pairs & 129 and 20 states \\
Sentences replace goal images & \Cref{tab:main} & Live, replay \\
Scores read the predictions & \Cref{tab:ablation} & Live, replay \\
\bottomrule
\end{tabular}
\end{table}

\subsection{Evaluation design}
\label{sec:setup}
\paragraph{Layouts and outcomes.}
We use LIBERO-Object scenes~\cite{liu2023libero} in robosuite~\cite{zhu2020robosuite}. From demonstration starts with an object already held, we shift the basket and place scene objects along the route to it. Development uses 15 layouts, and evaluation uses 88 generated with different seeds from 26 demonstrations across 10 tasks. Route obstacles appear in 56 evaluation layouts. The other 32 comprise direct layouts and short layouts, in which the basket starts close to the held object. Predictor and head training use 420 other demonstrations and 50 for validation, so evaluation starts are unseen while objects are shared. Of the evaluation layouts, 35 share a source demonstration with development layouts (supplementary Sec.~I), and we also report the other 53. Trials use planner seeds 1 to 3, giving 264 trials per evaluation cell. An event occurs when another object, including the basket, moves more than 10\,mm or rotates more than $15^\circ$, measured after each decision. Safe placement is completion without an event.

\paragraph{Live and replay.}
Live runs query GPT-6 Luna (\texttt{gpt-6-luna}) on their own current frames, with prompts fixed on development layouts. Replay uses the responses stored by the live sentence run on the same layout and seed, taking the nearest step with the same phase, so that cells differing in readout share their language input. Image-goal runs use language only for screening.

\paragraph{Selection and statistics.}
Each readout except raw staged images selects its weight $w$ by the same rule on development layouts. Raw cosine selects the same action as raw squared distance at every step. Projected final and staged images select 1.0 and 0.6, and sentences 0.6. A weight carried over from another readout or head without this selection is called inherited. Trials are paired by layout and seed, and exact two-sided sign tests compare layout totals over the three seeds. Intervals are 95\% and resample layouts. Where noted, they resample the 26 source demonstrations.

\paragraph{Executed-candidate probes.}
We restore a decision state and execute sampled candidates separately. Scores use pre-execution predictions, and labels use resulting simulator states, so all readouts in a probe see identical predictions. AUROC is computed within each state containing both outcomes and averaged across states.

\subsection{Progress toward the release point and the readout}
\label{sec:progress}
\begin{table}[t]
\centering
\caption{Scores added to progress in \cref{eq:select}, original head, 264 trials per cell. Final goals use the last frame of a successful execution in the evaluation layout. Staged goals use its middle frame in the far phase and its last frame in the near phase. Raw staged goals inherit the raw-final weight. Cells marked n/a were not run.}
\label{tab:controlled}
\footnotesize
\setlength{\tabcolsep}{2.5pt}
\begin{tabular}{@{}llrrr@{}}
\toprule
& & Live & \multicolumn{2}{c}{Replay} \\
Score added to progress & Goal input & Safe & Safe & Events \\
\midrule
None (head reads current frame) & n/a & n/a & 0 & 264 \\
Raw final image & Final frame & 94 & 97 & 163 \\
Raw staged images & Staged frames & n/a & 158 & 64 \\
Projected final image & Final frame & 185 & 168 & 58 \\
Projected staged images & Staged frames & 228 & 228 & 35 \\
Sentences & Current frame & 222 & 214 & 46 \\
\bottomrule
\end{tabular}
\end{table}

\paragraph{Distance to the release point alone does not place objects safely.}
When the head reads the current frame, the target and screening scores are equal for every candidate, and selection follows progress alone. Under replay, this planner completes 30 of 264 placements, and every trial has an event (\cref{tab:controlled}).

\begin{figure*}[t]
\centering
\includegraphics[width=\textwidth]{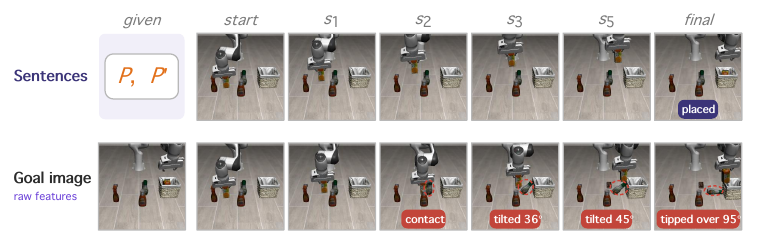}
\caption{Two replayed trials on one layout with planner seed 1, sharing the predictor, sampler, screening and weight. The top trial reads predictions with target sentences, and the bottom trial with raw-feature L1 distance to the goal image at left, taken from a safe execution in this layout. The two trials diverge after the first decision.}
\label{fig:qual}
\end{figure*}

\paragraph{Each readout adds to progress by ranking predicted arrangements.}
The rows of \cref{tab:controlled} differ only in the second score of \cref{eq:select}, its goal input and its selected weight. The controller, sampler and release rule that give no safe placement with progress alone give 97 with raw final images, 168 with projected final images and 214 with sentences (\cref{fig:qual}).

\subsection{The same goal image in two spaces}
\label{sec:controlled}

\paragraph{Projection changes closed-loop outcomes with the goal pixels fixed.}
In live runs, projection raises safe placement from 35.6\% to 70.1\% and reduces events from 167 to 49. It gives more safe placements in 58 layouts and fewer in 12 ($p=2\times10^{-8}$), a difference of 34.5 points. The two runs share the goal pixels, predictor, release point, screening and release rule, while the score, goal encoding and development-selected weight change. On the 53 layouts whose source demonstrations are absent from development, the rates are 42.1\% and 64.2\%. Split by layout type (\cref{tab:types}), projection gives more safe placements than raw features in six of the seven types, and the two tie at 8 of 21 in low-to-high layouts. Short layouts, in which the basket starts close to the held object, give sentences their lowest safe-placement rate.

\begin{table}[t]
\centering
\caption{Three head-training seeds, safe placements out of 264 replay trials. Raw final images, whose goal score does not use the head, give 97. Inherited weights, those selected for the original head, are 1.0 for images and 0.6 for sentences. Retuning uses development layouts only. Both retrained image heads select 0.6, and sentence heads select 0.6 and 0.8.}
\label{tab:heads}
\footnotesize
\setlength{\tabcolsep}{4pt}
\begin{tabular}{@{}lccc@{}}
\toprule
Projected readout, weight & Original & Seed 1 & Seed 2 \\
\midrule
Final image, inherited $w$ & 168 & 134 & 98 \\
Final image, retuned $w$ & 168 & 202 & 137 \\
Sentences, inherited $w$ & 214 & 210 & 218 \\
Sentences, retuned $w$ & 214 & 210 & 203 \\
\bottomrule
\end{tabular}
\end{table}

\begin{table}[t]
\centering
\caption{Live safe placements by layout type with simulator geometry and the original head, with the types whose route to the basket has an obstacle listed first.}
\label{tab:types}
\footnotesize
\setlength{\tabcolsep}{3pt}
\begin{tabular}{@{}lrrrrr@{}}
\toprule
& & \multicolumn{2}{c}{Final image} & & Direct \\
Layout type & Trials & Raw & Proj. & Sentences & VLM \\
\midrule
Middle & 57 & 18 & 48 & 53 & 27 \\
End & 36 & 14 & 20 & 33 & 20 \\
Start & 36 & 12 & 28 & 29 & 9 \\
Two, low to high & 21 & 8 & 8 & 17 & 6 \\
Two, high to low & 18 & 1 & 10 & 14 & 9 \\
\midrule
Direct & 57 & 35 & 46 & 54 & 28 \\
Short & 39 & 6 & 25 & 22 & 2 \\
\midrule
All & 264 & 94 & 185 & 222 & 101 \\
\bottomrule
\end{tabular}
\end{table}

\paragraph{The size of the change depends on the head.}
\Cref{tab:heads} adds two heads trained from other seeds with the same data and settings. With weights retuned on development layouts, projected final images give 168, 202 and 137 safe placements, and raw L1 gives 97. At the inherited weight, the seed-2 head gives 98. The live runs and staged-image experiments use the original head, so 70.1\% describes that head. Sentence scores through the same three heads give 214, 210 and 203 with retuned weights.

\paragraph{Staging and projection act on different parts of the image query.}
Under replay, staging raises raw final images from 97 to 158 safe placements and projected ones from 168 to 228. Staging changes which frame is queried, and projection changes how it is read. With staged goals, projection adds 26.5 points. In the live projected-final run, 30 trials end with neither an event nor completion. All exhaust the 16-decision budget, and 23 remain in the far phase. The final frame leaves the intermediate arrangement open, which motivates comparing sentences with staged images in \cref{sec:main}.

\subsection{Why whole-frame raw distance reads less}
\label{sec:space}
The projected readout differs from raw L1 in weight, metric, pooling and the learned map. The controls below change these one at a time, and the executed-candidate probes locate where the raw score loses information.

\paragraph{A shared weight retains the difference.}
Over five development weights, raw L1, squared and pooled distances reach at most 16, 16 and 9 of 45 safe placements, and the projected score 29 (supplementary Sec.~G). At the shared weight $w=0.6$ on evaluation layouts, projection gives 173 safe placements and raw L1 gives 97.

\begin{table}[t]
\centering
\caption{Metric and pooling controls over 264 replay trials. The two projected rows share the same score and differ in its planning weight.}
\label{tab:outcomes}
\footnotesize
\setlength{\tabcolsep}{3pt}
\begin{tabular}{@{}lrrr@{}}
\toprule
Final-image readout & $w$ & Safe & Events \\
\midrule
Raw patches, L1 & 0.6 & 97 & 163 \\
Raw patches, cosine & 0.6 & 99 & 164 \\
Raw patches, squared & 0.6 & 99 & 164 \\
Raw mean, L1 & 0.6 & 47 & 216 \\
Projected mean & 1.0 & 168 & 58 \\
Projected mean & 0.6 & 173 & 91 \\
\bottomrule
\end{tabular}
\end{table}

\paragraph{Changing the raw metric or pooling does not close the gap.}
Patchwise cosine and squared distance each give 99 safe placements, compared with 97 for L1 (\cref{tab:outcomes}). Averaging raw features before L1 gives 47. Applying the learned map to that same mean representation gives 168. The gain therefore comes with the map, and pooling alone lowers the raw score.

\paragraph{Local differences are present in the raw features.}
Obstacles occupy 2 to 13 of 256 feature patches in the probe states, with a median of about 7. On 33 states with mixed contact outcomes, whole-frame L1 ranks candidates for contact with AUROC 0.572. Restricting the distance to obstacle patches, located with the simulator, gives 0.685 [0.599, 0.769], higher in 23 states ($p=0.035$). This measurement locates information and does not enter planning. The raw features thus carry contact-related differences that the whole-frame distance dilutes.

\paragraph{The tested intervention is the learned map into SigLIP.}
All heads we trained map into SigLIP, because the second question needs images and sentences in one space. The target geometry and the alignment objective therefore vary together in these comparisons.

\subsection{Sentence scores against executed outcomes}
\label{sec:probe}
Before planning with sentences, we ask whether a sentence score computed from a prediction corresponds to the relation reached by executing that candidate. Locked phrases separate choosing the wording from evaluating it, and a second check uses the sentences written during planning.

\paragraph{Locked phrases transfer from frames to predictions.}
We choose a phrase pair on real future frames from a calibration subset of layouts, lock the wording, and evaluate predictions on new states excluded from phrase selection, with the head fixed. The pair contrasts an object held ``far above the basket'' with one ``at the bottom of the basket''. On 129 states from 76 layouts, it separates the executed basket relations with AUROC 0.695 [0.631, 0.784], above 0.5 in 93 states ($p=3\times10^{-7}$). On 36 near-phase states, sentence scores from predictions and from the actual future frames have mean within-state rank correlation $+0.35$ [0.26, 0.43], positive in 31 states. A second locked pair, the held object above or below a route obstacle, gives AUROC 0.618 [0.514, 0.758] on 42 states from 29 layouts ($p=0.044$).

\begin{figure}[t]
\centering
\includegraphics[width=\linewidth]{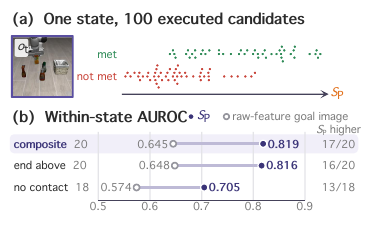}
\caption{Sentence scores checked against executed outcomes. (a) The 100 candidates of one far-phase state, placed by $S_P$ and colored by whether the executed motion meets the route relation. (b) Within-state AUROC for this relation and its two parts, with raw-feature goal-image L1 on the same predictions as open points.}
\label{fig:auroc}
\end{figure}

\paragraph{Generated sentences on a fixed route-relation test.}
We sample an event-free far-phase decision from each of 40 evaluation layouts with route obstacles in live seed-1 runs, draw 100 eligible candidates and execute each from the restored state with the closed-loop controller. The label combines the end-state relation to every route obstacle that the far-phase prompt requests with absence of contact along the motion (geometric predicate in supplementary Sec.~B). It is shared across queries and computed independently of the scores. On the 20 states with both outcomes, the VLM's target-contrast score gives mean AUROC 0.82 [0.75, 0.88] (\cref{fig:auroc}), with a per-state median of 0.873 and 18 states above 0.5. The fraction of candidates that meet the relation ranges from 0.01 to 0.98 across these states.

\paragraph{End state and intervening contact differ.}
Splitting the route label gives AUROC 0.816 for the end-state relation on 20 states and 0.705 for absence of route-obstacle contact on 18 states. The composite score mostly tracks the end-state relation. A sentence score thus measures compatibility of a predicted frame with a described arrangement more reliably than it measures contact along the executed motion.

\subsection{Replacing goal images with sentences}
\label{sec:main}
\begin{table}[t]
\centering
\caption{Replacing the image query while retaining the release point, 264 trials per row, original head. Same-layout images require a prior scripted successful execution in the evaluation layout. Source-demo images use the layout's source demonstration. The last row estimates both geometric points from RGB-D. Compl. counts completions with or without an event.}
\label{tab:main}
\footnotesize
\setlength{\tabcolsep}{2pt}
\begin{tabular}{@{}llrrr@{}}
\toprule
Goal query & Requires & Safe & Compl. & Events \\
\midrule
\multicolumn{5}{@{}l}{\emph{Live, simulator geometry}} \\
Staged images, same layout & Layout run & 228 & 258 & 36 \\
Sentences & Current frame & 222 & 249 & 42 \\
\midrule
\multicolumn{5}{@{}l}{\emph{Replay, simulator geometry}} \\
Staged images, same layout & Layout run & 228 & 257 & 35 \\
Staged images, source demo & Demonstration & 200 & 242 & 62 \\
Sentences & Current frame & 214 & 242 & 46 \\
\midrule
\multicolumn{5}{@{}l}{\emph{Live, RGB-D geometry}} \\
Sentences & Current frame & 198 & 232 & 53 \\
\bottomrule
\end{tabular}
\end{table}

\paragraph{Sentences and staged images give similar observed placement rates.}
Staged images and sentences switch at the same 12\,cm phase boundary and share the release point, sampler and screening, so this comparison keeps the staging and changes the query and its source. With simulator geometry, live sentences give 84.1\% safe placement and staged image goals 86.4\% (\cref{tab:main}). The sentence-minus-image difference is $-2.3$ points [$-8.7$, 4.2], with sentences higher in 12 layouts and images in 19 ($p=0.28$). The interval admits a deficit of up to 8.7 points and an advantage of up to 4.2.

\paragraph{The image goals depend on how they are obtained.}
Same-layout staged images show the rearranged scene after a successful execution. Each layout's source demonstration has the basket at its original position and lacks the added obstacles. Its middle and final frames, rendered in the evaluation environment, give 200 safe placements under replay, $-5.3$ points [$-13.6$, 3.0] relative to sentences ($p=0.41$). For same-layout images, taking the far-phase frame at a quarter, half or three quarters of the execution gives 189, 228 and 226 safe placements at the same weight.

\paragraph{RGB-D can supply the release point.}
Estimating the destination and held-object reference point gives 198 safe placements (75.0\%). Relative to simulator geometry, safe placements decrease in 26 layouts and increase in 8 ($p=0.003$), and resampling the 26 source demonstrations gives $-9.1$ points [$-16.0$, $-2.0$]. The perceived target differs from the simulator target by more than 2\,cm in 24 trials. In both variants the release point specifies where the object should end, and the sentence replaces the image as the query for the rest of the goal.

\subsection{Do the sentence scores read the predictions?}
\label{sec:ablation}
\begin{table}[t]
\centering
\caption{Where the selection information comes from, with simulator geometry. Each block includes its reference run. Counts are over 264 trials on evaluation layouts and 45 on development layouts. Direct control receives the release point and decides release itself. Columns follow \cref{tab:main}.}
\label{tab:ablation}
\footnotesize
\setlength{\tabcolsep}{3pt}
\begin{tabular}{@{}lrrr@{}}
\toprule
Intervention & Safe & Compl. & Events \\
\midrule
\multicolumn{4}{@{}l}{\emph{88 layouts, replay}} \\
Sentence readout & 214 & 242 & 46 \\
$P$ alone & 195 & 233 & 67 \\
$P$ minus neutral sentence & 101 & 150 & 159 \\
\midrule
\multicolumn{4}{@{}l}{\emph{88 layouts, live}} \\
Sentence readout & 222 & 249 & 42 \\
Same VLM selects motions directly & 101 & 139 & 149 \\
\midrule
\multicolumn{4}{@{}l}{\emph{15 development layouts, replay}} \\
Sentence readout & 35 & 42 & 9 \\
No target score ($w=0$) & 2 & 9 & 43 \\
$S_P$, $S_N$ shuffled & 5 & 11 & 40 \\
\midrule
\multicolumn{4}{@{}l}{\emph{15 development layouts, live}} \\
Sentence readout & 35 & 44 & 10 \\
VLM describes current frame & 11 & 25 & 34 \\
\bottomrule
\end{tabular}
\end{table}

\paragraph{Each score must come from its own candidate's prediction.}
On development layouts, removing the target score leaves screening and progress and gives 2 safe placements, and shuffling scores among candidates gives 5 (\cref{tab:ablation}). Keeping the predictions but asking the VLM to describe the observed frame lowers safe placements from 35 to 11. Summed over seeds per layout, each of these three changes gives fewer safe placements in 11 to 13 of the 15 layouts and more in none ($p\le0.001$). The selection information therefore comes from comparing each prediction with a described successor relation.

\paragraph{The same VLM without the predictions.}
Given the current frame, end-effector position, object names and the simulator release point, the same VLM returns the next position and a release decision under the same motion limit and decision budget. Used this way, it gives 101 safe placements (38.3\%), and reading the predictions with its sentences gives 222. Supplementary Sec.~G describes its inputs and a VoxPoser run with the same target point.

\paragraph{A released predictor.}
With the released V-JEPA 2-AC predictor, not fine-tuned and read through a head trained on its own features in the same way, sentences give 27 safe placements in 45 development trials under replay, with 18 events, compared with 35 and 9 with the fine-tuned predictor.

\paragraph{Choice of contrast.}
Using $P$ alone gives 195 safe placements, compared with 214 for $P-P'$ ($p=0.20$), and 8.0 points more events [0.8, 14.8]. Subtracting ``A photo of the [object]'' gives 101.

\section{Conclusion}
\label{sec:conclusion}
We asked whether a goal-image planner that disturbs the scene lacks the change in its predictions or hides it in its distance. For a frozen V-JEPA 2-AC predictor, distance to an exactly known release point places no object safely, and safe placement depends on how the predictions are read. The raw features hold contact-related differences that whole-frame distance dilutes, and the same goal image read through a learned projection into SigLIP gives more safe placements for each of three heads. In that space, sentence scores track tested relations on executed candidates, and VLM-written relations reach a placement rate close to that of staged images without any goal image. With the predictor fixed, the readout moved live safe placement by 34.5 points, so goal-image latent planners can be reported with their readout, its weight and, for learned readouts, several training seeds alongside the predictor.

\paragraph{Limitations.}
The study covers one simulated placement family, seen objects, one camera and single-step translations. All planners share the release point, phase and release rule, so the findings concern ranking predictions under a known destination. Pushing, rotation, multi-step search such as CEM, new objects and real robots are outside this study. Sentence scores depend on phrasing and on one closed-source VLM. Events are measured at decision boundaries, so transient contacts are not counted. On a physical robot, a VLM that misdescribes the scene would give a wrong target relation for that decision.
\label{lastbodypage}
{
\small
\bibliographystyle{ieeenat_fullname}
\bibliography{main1}

\begin{thebibliography}{45}
\providecommand{\natexlab}[1]{#1}
\providecommand{\url}[1]{\texttt{#1}}
\expandafter\ifx\csname urlstyle\endcsname\relax
  \providecommand{\doi}[1]{doi: #1}\else
  \providecommand{\doi}{doi: \begingroup \urlstyle{rm}\Url}\fi

\bibitem[Assran et~al.(2023)Assran, Duval, Misra, Bojanowski, Vincent, Rabbat,
  LeCun, and Ballas]{assran2023ijepa}
Mahmoud Assran, Quentin Duval, Ishan Misra, Piotr Bojanowski, Pascal Vincent,
  Michael Rabbat, Yann LeCun, and Nicolas Ballas.
\newblock Self-supervised learning from images with a joint-embedding
  predictive architecture.
\newblock In \emph{IEEE/CVF Conference on Computer Vision and Pattern
  Recognition (CVPR)}, 2023.

\bibitem[Assran et~al.(2025)Assran, Bardes, Fan, Garrido, Howes, Komeili,
  Muckley, Rizvi, Roberts, et~al.]{assran2025vjepa2}
Mido Assran, Adrien Bardes, David Fan, Quentin Garrido, Russell Howes, Mojtaba
  Komeili, Matthew~J. Muckley, Ammar Rizvi, Claire Roberts, et~al.
\newblock {V-JEPA} 2: Self-supervised video models enable understanding,
  prediction and planning.
\newblock \emph{arXiv preprint arXiv:2506.09985}, 2025.

\bibitem[Barbeau et~al.(2026)Barbeau, Roy, Beltrame, Desrosiers, and
  Thome]{barbeau2026lago}
Samuel Barbeau, Simon Roy, Giovanni Beltrame, Christian Desrosiers, and Nicolas
  Thome.
\newblock Latent goal prediction from language for model-based planning.
\newblock \emph{arXiv preprint arXiv:2606.20627}, 2026.

\bibitem[Bardes et~al.(2024)Bardes, Garrido, Ponce, Chen, Rabbat, LeCun,
  Assran, and Ballas]{bardes2024vjepa}
Adrien Bardes, Quentin Garrido, Jean Ponce, Xinlei Chen, Michael Rabbat, Yann
  LeCun, Mahmoud Assran, and Nicolas Ballas.
\newblock Revisiting feature prediction for learning visual representations
  from video.
\newblock \emph{Transactions on Machine Learning Research (TMLR)}, 2024.

\bibitem[Berg et~al.(2025)Berg, Zhu, Bao, Durugkar, and Gupta]{berg2025swm}
Jacob Berg, Chuning Zhu, Yanda Bao, Ishan Durugkar, and Abhishek Gupta.
\newblock Semantic world models.
\newblock \emph{arXiv preprint arXiv:2510.19818}, 2025.

\bibitem[Chua et~al.(2018)Chua, Calandra, McAllister, and Levine]{chua2018pets}
Kurtland Chua, Roberto Calandra, Rowan McAllister, and Sergey Levine.
\newblock Deep reinforcement learning in a handful of trials using
  probabilistic dynamics models.
\newblock In \emph{Advances in Neural Information Processing Systems
  (NeurIPS)}, 2018.

\bibitem[Du et~al.(2023{\natexlab{a}})Du, Konyushkova, Denil, Raju, Landon,
  Hill, de~Freitas, and Cabi]{du2023success}
Yuqing Du, Ksenia Konyushkova, Misha Denil, Akhil Raju, Jessica Landon, Felix
  Hill, Nando de Freitas, and Serkan Cabi.
\newblock Vision-language models as success detectors.
\newblock In \emph{Conference on Lifelong Learning Agents (CoLLAs)},
  2023{\natexlab{a}}.

\bibitem[Du et~al.(2023{\natexlab{b}})Du, Yang, Dai, Dai, Nachum, Tenenbaum,
  Schuurmans, and Abbeel]{du2023unipi}
Yilun Du, Mengjiao Yang, Bo Dai, Hanjun Dai, Ofir Nachum, Joshua~B. Tenenbaum,
  Dale Schuurmans, and Pieter Abbeel.
\newblock Learning universal policies via text-guided video generation.
\newblock In \emph{Advances in Neural Information Processing Systems
  (NeurIPS)}, 2023{\natexlab{b}}.

\bibitem[Du et~al.(2024)Du, Yang, Florence, Xia, Wahid, Ichter, Sermanet, Yu,
  Abbeel, Tenenbaum, et~al.]{du2024vlp}
Yilun Du, Mengjiao Yang, Pete Florence, Fei Xia, Ayzaan Wahid, Brian Ichter,
  Pierre Sermanet, Tianhe Yu, Pieter Abbeel, Joshua~B. Tenenbaum, et~al.
\newblock Video language planning.
\newblock In \emph{International Conference on Learning Representations
  (ICLR)}, 2024.

\bibitem[Ebert et~al.(2018)Ebert, Finn, Dasari, Xie, Lee, and
  Levine]{ebert2018foresight}
Frederik Ebert, Chelsea Finn, Sudeep Dasari, Annie Xie, Alex~X. Lee, and Sergey
  Levine.
\newblock Visual foresight: Model-based deep reinforcement learning for
  vision-based robotic control.
\newblock \emph{arXiv preprint arXiv:1812.00568}, 2018.

\bibitem[Fang et~al.(2024)Fang, Liu, Abbeel, and Levine]{fang2024moka}
Kuan Fang, Fangchen Liu, Pieter Abbeel, and Sergey Levine.
\newblock {MOKA}: Open-world robotic manipulation through mark-based visual
  prompting.
\newblock In \emph{Robotics: Science and Systems (RSS)}, 2024.

\bibitem[Finn and Levine(2017)]{finn2017foresight}
Chelsea Finn and Sergey Levine.
\newblock Deep visual foresight for planning robot motion.
\newblock In \emph{IEEE International Conference on Robotics and Automation
  (ICRA)}, 2017.

\bibitem[Goswami et~al.(2025)Goswami, Bar, Fan, Yang, Zhou, Krishnamurthy,
  Rabbat, Khorrami, and LeCun]{goswami2025dexwm}
Raktim~Gautam Goswami, Amir Bar, David Fan, Tsung-Yen Yang, Gaoyue Zhou,
  Prashanth Krishnamurthy, Michael Rabbat, Farshad Khorrami, and Yann LeCun.
\newblock World models for learning dexterous hand-object interactions from
  human videos.
\newblock \emph{arXiv preprint arXiv:2512.13644}, 2025.

\bibitem[Ha and Schmidhuber(2018)]{ha2018worldmodels}
David Ha and Jürgen Schmidhuber.
\newblock Recurrent world models facilitate policy evolution.
\newblock In \emph{Advances in Neural Information Processing Systems
  (NeurIPS)}, 2018.

\bibitem[Hafner et~al.(2019)Hafner, Lillicrap, Fischer, Villegas, Ha, Lee, and
  Davidson]{hafner2019planet}
Danijar Hafner, Timothy Lillicrap, Ian Fischer, Ruben Villegas, David Ha,
  Honglak Lee, and James Davidson.
\newblock Learning latent dynamics for planning from pixels.
\newblock In \emph{International Conference on Machine Learning (ICML)}, 2019.

\bibitem[Hafner et~al.(2020)Hafner, Lillicrap, Ba, and
  Norouzi]{hafner2020dreamer}
Danijar Hafner, Timothy Lillicrap, Jimmy~Lei Ba, and Mohammad Norouzi.
\newblock Dream to control: Learning behaviors by latent imagination.
\newblock In \emph{International Conference on Learning Representations
  (ICLR)}, 2020.

\bibitem[Hafner et~al.(2025)Hafner, Pašukonis, Ba, and
  Lillicrap]{hafner2025dreamerv3}
Danijar Hafner, Jurgis Pašukonis, Jimmy~Lei Ba, and Timothy Lillicrap.
\newblock Mastering diverse control tasks through world models.
\newblock \emph{Nature}, 640:\penalty0 647--653, 2025.

\bibitem[Hansen et~al.(2022)Hansen, Wang, and Su]{hansen2022tdmpc}
Nicklas Hansen, Xiaolong Wang, and Hao Su.
\newblock Temporal difference learning for model predictive control.
\newblock In \emph{International Conference on Machine Learning (ICML)}, 2022.

\bibitem[Hansen et~al.(2024)Hansen, Su, and Wang]{hansen2024tdmpc2}
Nicklas Hansen, Hao Su, and Xiaolong Wang.
\newblock {TD-MPC2}: Scalable, robust world models for continuous control.
\newblock In \emph{International Conference on Learning Representations
  (ICLR)}, 2024.

\bibitem[Hua et~al.(2026)Hua, Das, Boominathan, and Balestriero]{hua2026levlwm}
Qiuyi Hua, Ahitagni Das, Vivek Boominathan, and Randall Balestriero.
\newblock {LeVLWM: Text-Goal Control with a Single Vision-Language World
  Model}.
\newblock In \emph{NeurIPS Workshop on Neuro-Symbolic Embodied Intelligence},
  2026.

\bibitem[Huang et~al.(2023)Huang, Wang, Zhang, Li, Wu, and
  Fei-Fei]{huang2023voxposer}
Wenlong Huang, Chen Wang, Ruohan Zhang, Yunzhu Li, Jiajun Wu, and Li Fei-Fei.
\newblock {VoxPoser}: Composable {3D} value maps for robotic manipulation with
  language models.
\newblock In \emph{Conference on Robot Learning (CoRL)}, 2023.

\bibitem[Huang et~al.(2024)Huang, Wang, Li, Zhang, and Fei-Fei]{huang2024rekep}
Wenlong Huang, Chen Wang, Yunzhu Li, Ruohan Zhang, and Li Fei-Fei.
\newblock {ReKep}: Spatio-temporal reasoning of relational keypoint constraints
  for robotic manipulation.
\newblock In \emph{Conference on Robot Learning (CoRL)}, 2024.

\bibitem[Khatib(1987)]{khatib1987osc}
Oussama Khatib.
\newblock A unified approach for motion and force control of robot
  manipulators: The operational space formulation.
\newblock \emph{IEEE Journal on Robotics and Automation}, 3\penalty0
  (1):\penalty0 43--53, 1987.

\bibitem[Li et~al.(2026)Li, Feng, Zhang, Li, Wang, Alahi, and Soh]{li2026gwm}
Quanyi Li, Lan Feng, Haonan Zhang, Wuyang Li, Letian Wang, Alexandre Alahi, and
  Harold Soh.
\newblock Grounded world model: Latent planning with language goals.
\newblock \emph{arXiv preprint arXiv:2604.11751}, 2026.

\bibitem[Lin et~al.(2024)Lin, Du, Watkins, Hafner, Abbeel, Klein, and
  Dragan]{lin2024dynalang}
Jessy Lin, Yuqing Du, Olivia Watkins, Danijar Hafner, Pieter Abbeel, Dan Klein,
  and Anca~D. Dragan.
\newblock Learning to model the world with language.
\newblock In \emph{International Conference on Machine Learning (ICML)}, 2024.

\bibitem[Liu et~al.(2023)Liu, Zhu, Gao, Feng, Liu, Zhu, and
  Stone]{liu2023libero}
Bo Liu, Yifeng Zhu, Chongkai Gao, Yihao Feng, Qiang Liu, Yuke Zhu, and Peter
  Stone.
\newblock {LIBERO}: Benchmarking knowledge transfer for lifelong robot
  learning.
\newblock In \emph{Advances in Neural Information Processing Systems (NeurIPS)
  Datasets and Benchmarks Track}, 2023.

\bibitem[Ma et~al.(2023)Ma, Liang, Som, Kumar, Zhang, Bastani, and
  Jayaraman]{ma2023liv}
Yecheng~Jason Ma, William Liang, Vaidehi Som, Vikash Kumar, Amy Zhang, Osbert
  Bastani, and Dinesh Jayaraman.
\newblock {LIV}: Language-image representations and rewards for robotic
  control.
\newblock In \emph{International Conference on Machine Learning (ICML)}, 2023.

\bibitem[Mazzaglia et~al.(2024)Mazzaglia, Verbelen, Dhoedt, Courville, and
  Rajeswar]{mazzaglia2024genrl}
Pietro Mazzaglia, Tim Verbelen, Bart Dhoedt, Aaron~C. Courville, and Sai
  Rajeswar.
\newblock {GenRL}: Multimodal-foundation world models for generalization in
  embodied agents.
\newblock In \emph{Advances in Neural Information Processing Systems
  (NeurIPS)}, 2024.

\bibitem[Moayeri et~al.(2023)Moayeri, Rezaei, Sanjabi, and
  Feizi]{moayeri2023text2concept}
Mazda Moayeri, Keivan Rezaei, Maziar Sanjabi, and Soheil Feizi.
\newblock Text-to-concept (and back) via cross-model alignment.
\newblock In \emph{International Conference on Machine Learning (ICML)}, 2023.

\bibitem[Nair et~al.(2018)Nair, Pong, Dalal, Bahl, Lin, and
  Levine]{nair2018rig}
Ashvin Nair, Vitchyr~H. Pong, Murtaza Dalal, Shikhar Bahl, Steven Lin, and
  Sergey Levine.
\newblock Visual reinforcement learning with imagined goals.
\newblock In \emph{Advances in Neural Information Processing Systems
  (NeurIPS)}, 2018.

\bibitem[Nair et~al.(2021)Nair, Mitchell, Chen, Ichter, Savarese, and
  Finn]{nair2021lorel}
Suraj Nair, Eric Mitchell, Kevin Chen, Brian Ichter, Silvio Savarese, and
  Chelsea Finn.
\newblock Learning language-conditioned robot behavior from offline data and
  crowd-sourced annotation.
\newblock In \emph{Conference on Robot Learning (CoRL)}, 2021.

\bibitem[Nasiriany et~al.(2024)Nasiriany, Xia, Yu, et~al.]{nasiriany2024pivot}
Soroush Nasiriany, Fei Xia, Wenhao Yu, et~al.
\newblock {PIVOT}: Iterative visual prompting elicits actionable knowledge for
  {VLMs}.
\newblock In \emph{International Conference on Machine Learning (ICML)}, 2024.

\bibitem[Radford et~al.(2021)Radford, Kim, Hallacy, Ramesh, Goh, Agarwal,
  Sastry, Askell, Mishkin, Clark, et~al.]{radford2021clip}
Alec Radford, Jong~Wook Kim, Chris Hallacy, Aditya Ramesh, Gabriel Goh,
  Sandhini Agarwal, Girish Sastry, Amanda Askell, Pamela Mishkin, Jack Clark,
  et~al.
\newblock Learning transferable visual models from natural language
  supervision.
\newblock In \emph{International Conference on Machine Learning (ICML)}, 2021.

\bibitem[Rocamonde et~al.(2024)Rocamonde, Montesinos, Nava, Perez, and
  Lindner]{rocamonde2024vlmrm}
Juan Rocamonde, Victoriano Montesinos, Elvis Nava, Ethan Perez, and David
  Lindner.
\newblock Vision-language models are zero-shot reward models for reinforcement
  learning.
\newblock In \emph{International Conference on Learning Representations
  (ICLR)}, 2024.

\bibitem[Sobal et~al.(2025)Sobal, Zhang, Cho, Balestriero, Rudner, and
  LeCun]{sobal2025pldm}
Vlad Sobal, Wancong Zhang, Kyunghyun Cho, Randall Balestriero, Tim G.~J.
  Rudner, and Yann LeCun.
\newblock Learning from reward-free offline data: A case for planning with
  latent dynamics models.
\newblock In \emph{Advances in Neural Information Processing Systems
  (NeurIPS)}, 2025.

\bibitem[Sontakke et~al.(2023)Sontakke, Zhang, Arnold, Pertsch, Bıyık,
  Sadigh, Finn, and Itti]{sontakke2023roboclip}
Sumedh Sontakke, Jesse~M. Zhang, Sébastien M.~R. Arnold, Karl Pertsch, Erdem
  Bıyık, Dorsa Sadigh, Chelsea Finn, and Laurent Itti.
\newblock {RoboCLIP}: One demonstration is enough to learn robot policies.
\newblock In \emph{Advances in Neural Information Processing Systems
  (NeurIPS)}, 2023.

\bibitem[Terver et~al.(2026)Terver, Yang, Ponce, Bardes, and
  LeCun]{terver2025jepawms}
Basile Terver, Tsung-Yen Yang, Jean Ponce, Adrien Bardes, and Yann LeCun.
\newblock What drives success in physical planning with joint-embedding
  predictive world models?
\newblock \emph{Transactions on Machine Learning Research (TMLR)}, 2026.

\bibitem[van~den Oord et~al.(2018)van~den Oord, Li, and Vinyals]{oord2018cpc}
Aäron van~den Oord, Yazhe Li, and Oriol Vinyals.
\newblock Representation learning with contrastive predictive coding.
\newblock \emph{arXiv preprint arXiv:1807.03748}, 2018.

\bibitem[Wu et~al.(2025)Wu, Tian, Swamy, and Bajcsy]{wu2025forewarn}
Yilin Wu, Ran Tian, Gokul Swamy, and Andrea Bajcsy.
\newblock From foresight to forethought: {VLM}-in-the-loop policy steering via
  latent alignment.
\newblock In \emph{Robotics: Science and Systems (RSS)}, 2025.

\bibitem[Zhai et~al.(2022)Zhai, Wang, Mustafa, Steiner, Keysers, Kolesnikov,
  and Beyer]{zhai2022lit}
Xiaohua Zhai, Xiao Wang, Basil Mustafa, Andreas Steiner, Daniel Keysers,
  Alexander Kolesnikov, and Lucas Beyer.
\newblock {LiT}: Zero-shot transfer with locked-image text tuning.
\newblock In \emph{IEEE/CVF Conference on Computer Vision and Pattern
  Recognition (CVPR)}, 2022.

\bibitem[Zhai et~al.(2023)Zhai, Mustafa, Kolesnikov, and Beyer]{zhai2023siglip}
Xiaohua Zhai, Basil Mustafa, Alexander Kolesnikov, and Lucas Beyer.
\newblock Sigmoid loss for language image pre-training.
\newblock In \emph{IEEE/CVF International Conference on Computer Vision
  (ICCV)}, 2023.

\bibitem[Zhang et~al.(2026)Zhang, Terver, Zholus, Chitnis, Sutaria, Assran,
  Balestriero, Bar, LeCun, and Ballas]{zhang2026hwm}
Wancong Zhang, Basile Terver, Artem Zholus, Soham Chitnis, Harsh Sutaria, Mido
  Assran, Randall Balestriero, Amir Bar, Yann LeCun, and Nicolas Ballas.
\newblock Hierarchical planning with latent world models.
\newblock In \emph{Conference on Robot Learning (CoRL)}, 2026.

\bibitem[Zhao et~al.(2024)Zhao, Chen, et~al.]{zhao2024vlmpc}
Wentao Zhao, Jiaming Chen, et~al.
\newblock {VLMPC}: Vision-language model predictive control for robotic
  manipulation.
\newblock In \emph{Robotics: Science and Systems (RSS)}, 2024.

\bibitem[Zhou et~al.(2025)Zhou, Pan, LeCun, and Pinto]{zhou2025dinowm}
Gaoyue Zhou, Hengkai Pan, Yann LeCun, and Lerrel Pinto.
\newblock {DINO-WM}: World models on pre-trained visual features enable
  zero-shot planning.
\newblock In \emph{International Conference on Machine Learning (ICML)}, 2025.

\bibitem[Zhu et~al.(2020)Zhu, Wong, Mandlekar, Martín-Martín, Joshi, Lin,
  Maddukuri, Nasiriany, and Zhu]{zhu2020robosuite}
Yuke Zhu, Josiah Wong, Ajay Mandlekar, Roberto Martín-Martín, Abhishek Joshi,
  Kevin Lin, Abhiram Maddukuri, Soroush Nasiriany, and Yifeng Zhu.
\newblock robosuite: A modular simulation framework and benchmark for robot
  learning.
\newblock \emph{arXiv preprint arXiv:2009.12293}, 2020.

\end{thebibliography}
}
\clearpage
\includepdf[pages=-]{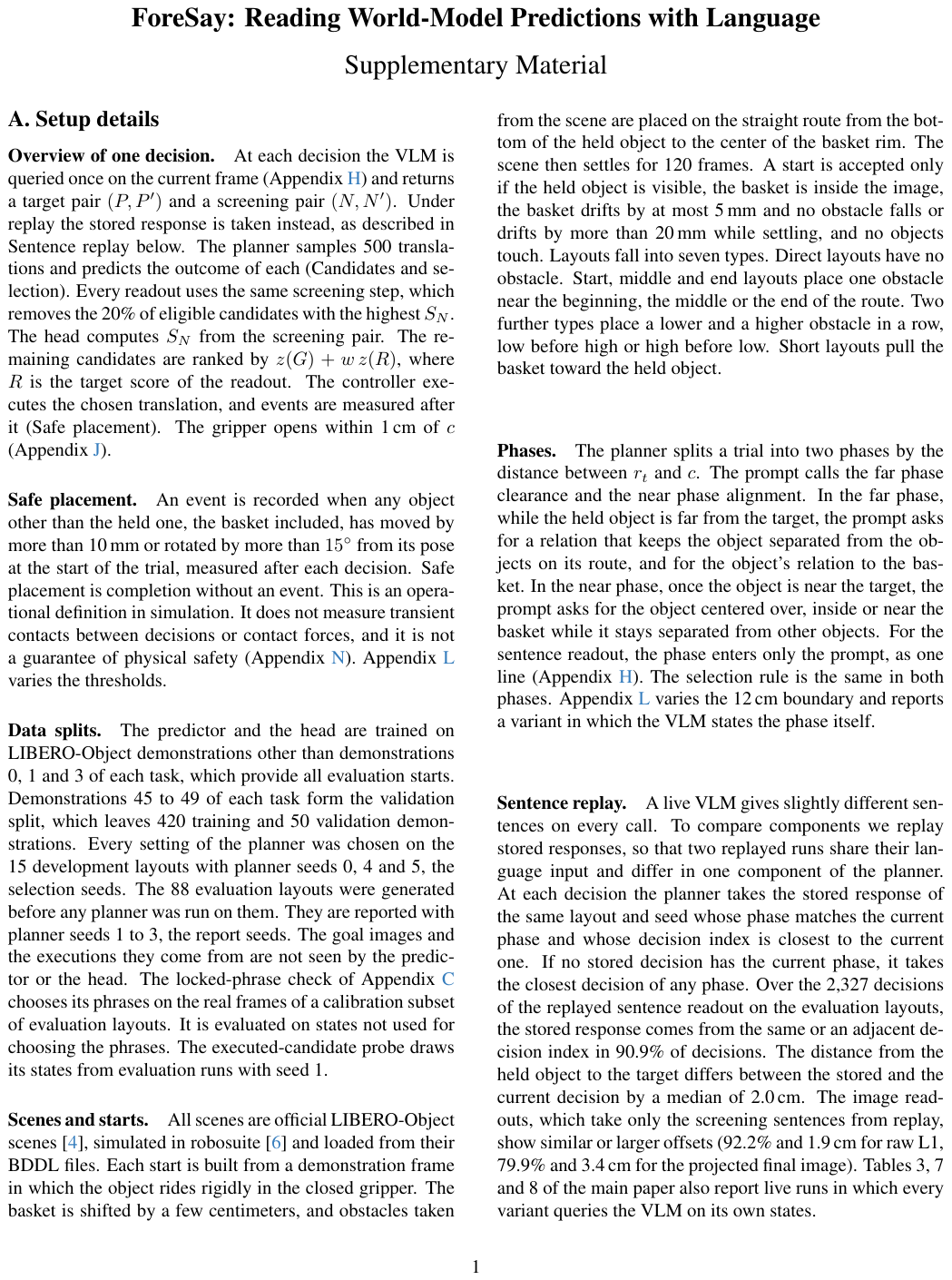}
\end{document}